\documentclass[10pt,twocolumn,letterpaper]{article}

\usepackage{cvpr}
\usepackage{times}
\usepackage{graphicx}
\usepackage{amsmath}
\usepackage{amssymb}
\usepackage{caption}
\usepackage{comment}
\usepackage{subcaption}
\usepackage{graphbox}
\usepackage{overpic}

\usepackage{caption}
\usepackage{graphbox} 
\usepackage{mwe}
\usepackage[subtle]{savetrees}
\usepackage[export]{adjustbox}

\usepackage[breaklinks=true,colorlinks,bookmarks=false]{hyperref}

\cvprfinalcopy 

\def\cvprPaperID{1}

\begin{document}

\title{Geometric Style Transfer\vspace{1em}\\%
    \includegraphics[width=\linewidth]{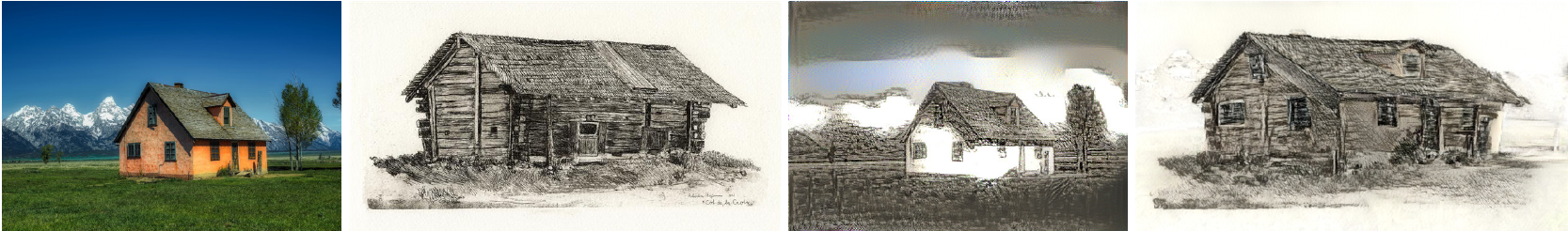}\\\captionof{figure}{Left to right: content image, style image, 
    texture only transfer using Gatys~\etal~\cite{gatys2016image}, 
    geometric and texture transfer using our method.
    Our method can not only capture texture features of the style image, 
    but also deform the content image to match geometric structures of the style image.}\label{fig:teaser}%
}

\author{Xiao-Chang Liu \\
University of Bath\\
\and
Xuan-Yi Li \quad Ming-Ming Cheng\\
Nankai University\\
\and
Peter Hall \\
University of Bath\\
}

\maketitle

%%%%%%%%% ABSTRACT
\begin{abstract}
   Neural style transfer (NST),
   where an input image is rendered in the style of another image,
   has been a topic of considerable progress in recent years. 
   Research over that time has been dominated by 
   transferring aspects of color and texture, yet
   these factors are only one component of style.
   Other factors of style include composition, the projection system used, and the way in which artists warp and bend objects.
   Our contribution is to introduce 
   a neural architecture that supports transfer of geometric style.
   Unlike recent work in this area, we are unique in being general in that we are not restricted by semantic content. This new architecture runs prior to a network that transfers texture style, enabling us to transfer texture to a warped image. This form of network supports a second novelty: we extend the NST input paradigm. Users can input content/style pair as is common, or they can chose to input a content/texture-style/geometry-style triple. This three image input paradigm divides style into two parts and so provides significantly greater versatility to the output we can produce. We provide user studies that show the quality of our output, and quantify the importance of geometric style transfer to style recognition by humans.
\end{abstract}

%-------------------------------------------------------------------------

\section{Introduction}

Neural style transfer (NST) is an active area of research with the  aim of synthesising artistic images. The most common paradigm is to input two images, one provides the content that the output should  contain, the other input indicates the ``style'' in which the provided content is to be rendered. To date, NST has been dominated by the transfer of texture, but artistic style is not characterised by texture alone. Style includes changes of shape of objects, rules of composition, the projection model used, and many other factors.

Our contribution is to step closer to artistic styles by including {\em geometric style transfer} (GST) that changes the shape of the content image to better match shapes in the style image. Figure~\ref{fig:teaser} demonstrates that transfer of geometric style yields an output that is a closer match to the style image that can be achieved by texture alone. This paper appeals to a little of the Art History literature to argue for the importance of GST, explains how it can be achieved in a general setting, and provides empirical evidence that GST yields output closer to the target style than texture transfer alone.

The style of an artist, or a school of artists, is only partly characterized by the way they make marks on the image surface. 
More than sixty years ago, Art Theorist Rudolf Arnheim~\cite{arnheim1954art}  argued that art style should be described by  attributes such as color, shape, and composition. Just over forty years later another Art Theorist, John Willats, was also concerned with art style and coined the terms {\em projection style} and {\em denotation style}. 
Denotation refers to the way in which an artist makes marks on the image surface. Denotation is impacted by the substrate (paper, canvas, {\em etc.}),    the media (paint, pencil, {\em etc.}) and the application method. Projection refers to the spatial organisation of parts~\cite{willats1997art}, it includes both standard cameras and orthogonal cameras, but more generally refers to the spatial organization of an objects parts. Willats shows that projection variety is at least if not more important than denotation variety in characterizing style. For example, ancient Egyptian art is characterised by the way people are unnaturally posed; Byzantine artists routinely used inverse perspective; Chinese artists traditionally used orthogonal projection;  Cubism exhibits a multitude of views in a single image. Denotation is important, but ancient Egyptian art (say) is recognizable in paintings, in bas-relief, in sculpture -- all of which differ in denotation but share the same ``projection'' variety.

It is worth noting that the geometric changes humans introduce tend not to be arbitrary. Some artists deform shape to bring emphasis to some aspect of the subject being depicted. For example, Stubbs would deliberately paint bulls to be larger and stronger than any real bull could be -- he did so to please the landowners whose animals he was depiciting. Other artists, such as Modigliani and El Greco, distort faces and humans as a matter of personal style. We can imagine that there is some underlying  photograph that has been somehow warped, and then painted over.

NST has been dominated by the transfer of texture, which approximates denotation.   There are some examples of {\em geometric style} transfer in NST,  but these require strong models of faces (\eg \cite{Yaniv2019,shi2019warpgan}) or of text \cite{Yang_textNST_2019_ICCV} so have a limited content domain. Our contribution is to introduce {\em geometric style} transfer as a general case to sit alongside texture transfer. Our method (see Section~\ref{sec:method}) conforms to the standard paradigm of providing a content and a style image, but in our case changes and distortions of geometry are transferred in addition to texture.  We can unique do style transfer with three inputs (content source, style source and geometric source). Processing the geometric changes requires an additional processing path that is parallel to an otherwise standard NST architecture.
The additional path is used to compute a geometric mapping that
warps the content image before it texture transfer takes place.
Section~\ref{sec:experimental-results} shows that our results not only tend to be preferred to the output of alternative algorithms, but also then to be regarded as more similar to the target output.

%-------------------------------------------------------------------------
\section{Related Work and Background}
\label{sec:background}

Image stylization is the process of mapping an input image 
in one style to an output image in a new style, with content preserved.   The problem has been a significant field of research within Visual Computing for over two decades, beginning with non-photorealistic rendering (NPR) and more recently continuing with neural style transfer (NST).

NPR has been the subject of research for many years and is too large to give a comprehensive overview. It includes image synthesis from 3D models, the emulation of substrate and media, and user interaction, but we confine ourselves to NPR from images. Early algorithms mapped pixel patches into ``blobs''~\cite{haeberli1990paint}.  Later, the mapping became more sophisticated, targeting salient regions~\eg~\cite{decarlo2002stylization}; 
blobs became brush strokes~\cite{hertzmann2003survey}.  These few examples are indicative of a much larger body of work in which image stylization is seen as a sophisticated filtering process. 
Higher forms of abstraction are far less common, but have been tackled using {\em ad-hoc} approaches emulating movements such as Cubism~\cite{collomosse2003cubist} and artists such as Archimboldo~\cite{huang2011arcimboldo}. 
Projection style in the sense Willats intends has been wholly neglected in NST, though there are examples in NPR such as~\cite{yu2004framework,hall2007rtcams}. 
Some of these early algorithms contained stochastic elements 
but were all prescriptive in the sense that the style of the output was predefined. Examples of learning style appeared as early as 1998~\cite{hall1998example}, and later in 2001~\cite{hertzmann2001image}.

Image stylization moved firmly in the direction of learning, in about 2015, when Gatys~\etal~\cite{Gatys2015c} introduced neural style transfer. The key idea was to adjust a variable image $X$ so that it matched some image $A$ for content and another image $B$ for style. The definition for content loss and style loss were both premised on features extracted from a network pre-trained for recognition (VGG-16 was used, \cite{simonyan2014very}),
the loss for content being the L2-norm between response vectors,
and the style loss being the L2-norm between Gram matrices comprising feature correlations.

The ability to learn style transforms is useful, 
but slow optimization motivated work towards fast transfer
~\cite{jing2018stroke,johnson2016perceptual,huang2017arbitrary, li2019learning}. 
A second problem is the need to retrain the network for each new style, 
which encouraged work to learn styles more generally, 
including but not limited to~\cite{li2017universal, xu2018learning,chen2017stylebank}. 
The loss functions have received attention, Huang~\etal~\cite{huang2017arbitrary} provide 
one example in which the loss function is based on the statistical distribution of features; 
Li~\etal~\cite{li2017universal} are another -- they use a whitening 
and coloring transforms to better map feature vectors. 
Kotovenko~\etal~\cite{Kotovenko_2019_ICCV} introduce two additional losses that learn subtle variations within one style and ensure stylization is not conditioned on the input photograph.

To date, NST has been extended to do many different tasks~\cite{jing2017neural}, 
such as portrait painting style transfer~\cite{selim2016painting,Yaniv2019};
visual attribute transfer~\cite{Liao:2017:VAT:3072959.3073683,kolkin2019style,yao2019attention,kotovenko2019content};
semantic style transfer in natural images~\cite{mechrez2018contextual,chen2016towards,champ2016semantic};
video style transfer~\cite{Ruder2018ArtisticST,huang2017real,gupta2017characterizing,chen2017coherent,li2019learning};
3D style transfer~\cite{chen2018stereoscopic,kato2018renderer},
and photorealistic style transfer~\cite{luan2017deep,mechrez2017photorealistic,li2018closed,Yoo_2019_ICCV}.

Nearly all NST is limited in the sense that there is no explicit attempt to change the geometry or shape of objects in the picture.
For clarity, many NST algorithms can output images with a different geometry to the input, but any such changes are accidents of the texture transfer process and are typically confined to the boundaries of objects. There is usually no effort to  deliberately transfer any geometric distortions introduced by an artist.
This limits the ability of NST algorithms to emulate style,
because geometric changes are an integral part of style.

The need to transfer {\em geometric} style has been recognised within NST, albeit in specialised domains.  Facial caricature is relatively popular ~\cite{yuphoto,li2018carigan,cao2018carigans,shi2019warpgan}, while Yaniv~\etal~\cite{Yaniv2019} consider artistic portraiture more generally; all of that work is limited to faces. Yang~\etal~\cite{Yang_textNST_2019_ICCV} explicitly  control the shape deformation of artistic text. Our work is unique by being the first to provide a general approach to geometric style transfer.

In summary, NST research has largely followed the trajectory of NPR in that work began on texture and later moved towards high forms of abstraction. If the history of NPR is a teacher, then NST will continue to develop away from texture transfer.

\section{Texture and Geometric Style Transfer}
\label{sec:method}

Our system transfers both texture and geometric styles, the latter being our contribution.   As is usual for neural style transfer, our system can take two inputs:  a content image $I^c$ and a style image $I^s$. Our novelty is this: rather than transfer texture directly, the content image is warped first. 
The warp carries the content image onto the style target, 
so the output image $I^o$ is the same size as $I^s$.
The warp transfers geometric style.  Figure~\ref{fig:ablation1} illustrates by showing photo-textured warped images and the textured results that come from them. 

Our system is novel too in  being able to accept three inputs: one content image as before, one geometric style image, and one texture style image. This paper is written assuming two inputs for familiarity and simplicity of explanation -- the extension to three inputs simply use the geometric style image through the geometric warping network, and the texture style image through the texture transfer network.

The whole procedure consists of three major steps: feature extraction, geometric warping, 
and texture transfer.
% as seen in Figure~\ref{fig:pipeline}. 
The first step is to extract features from a network, as explained in Section~\ref{sec:fea-extrac}. 
As is common in style transfer, we need ``content features'' to preserve content, 
``denotation'' features that will be used for denotation (texture) transfer, 
and uniquely we need  ``geometric features'' for geometry transfer. 
The second step sends these geometric features into a CNN architecture 
to compute a mapping $\Re^2 \mapsto \Re^2$; 
the mapping is used to warp the content image, see Section~\ref{sec:com-align}. 
The third step uses the content and texture features in an Image-Optimization-Based 
online style transfer method with a multi-scale strategy to transfer texture 
while preserving content to generate the final result, 
as described in Section~\ref{sec:style-trans}.

\section{Feature Extraction}\label{sec:fea-extrac}

\begin{figure}[tbp]
	\centering
	\includegraphics[width=\linewidth]{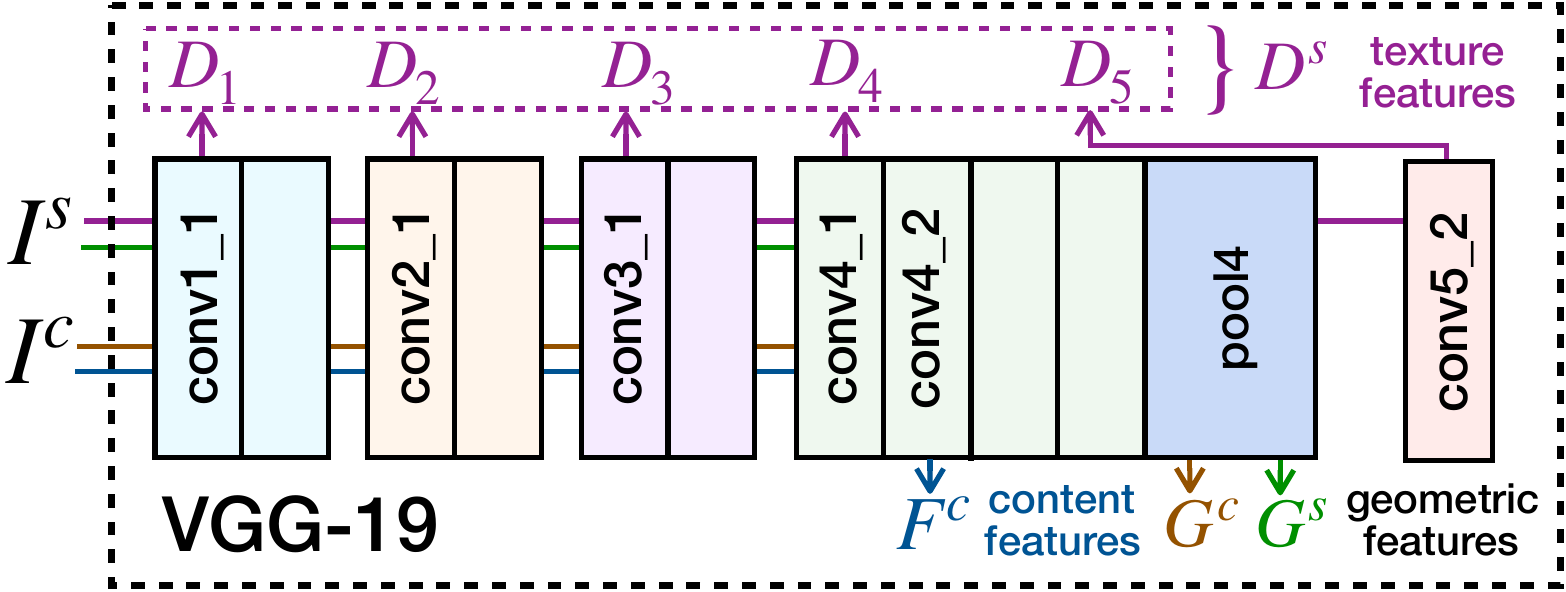}
	\caption{Feature extraction.}
	\label{fig:feature}
\end{figure}

We extract all our features from VGG-19,
which is trained on more than a million images from the ImageNet dataset~\cite{deng2009imagenet} 
and can classify images into a thousand or more object categories.  
A given input image is encoded in each layer by the filter responses to that image,
and layer $l$ with $N_l$ filters has $N_l$ distinct feature maps each of size $W_l\times{H_l}$.
%The network contains multi-level hierarchical features
The extraction procedure is summarised in Figure~\ref{fig:feature}.

\smallskip
\noindent{\bf Content Features } 
are used to preserve content during transfer. 
We draw them from the intermediate layer of the network, 
because such layers contain mid to high level image representations. 
Specifically, we use the $N_4$-element feature vectors from a $W_4
\times H_4$ array at at layer $conv4\_2$
to obtain feature maps  $F^c$ for the content image $I^c$ and $F^o$ for the output image.
These maps are each of size $W_4 \times H_4 \times N_4$,

\smallskip
\noindent{\bf Texture Features } 
are used to transfer denotation style 
(which prior work refers to as ``style'', with no modifying adjective). 
Low-level statistics tend to characterize denotation, but these benefit from context.
Following~\cite{gatys2016image,johnson2016perceptual},
we use early $conv1\_1$, $conv2\_1$, and later layer $conv3\_1$, $conv4\_1$ and $conv5\_1$,
and compute feature correlations as Gram matrices $D_l\in\Re^{N_l\times{N_l}}$ for each layer $l$.
Taken over all layers we have a set $D = \{ D_{l} : \forall l \in [1,2,3,4,5]\}$.
This set of correlations characterizes the texture style of an image;  $D^s$ for the style image $I^s$ and $D^o$ for the output image.

\smallskip
\noindent{\bf Geometric Features } 
are used to compute a spatial mapping $\mathcal{M}: \Re^2 \mapsto \Re^2$. The features should be reasonably robust to local local spatial variability in the form of rotation and translation, and aggregate content. 
We use the feature map at $pool4$ layer, followed by L2-normalization of each feature channel,
and get geometric features $G^c$ and $G^s$ for $I^c$ and $I^s$ respectively.

\begin{figure*}[tbp]
	\centering
    \includegraphics[width=0.75\linewidth]{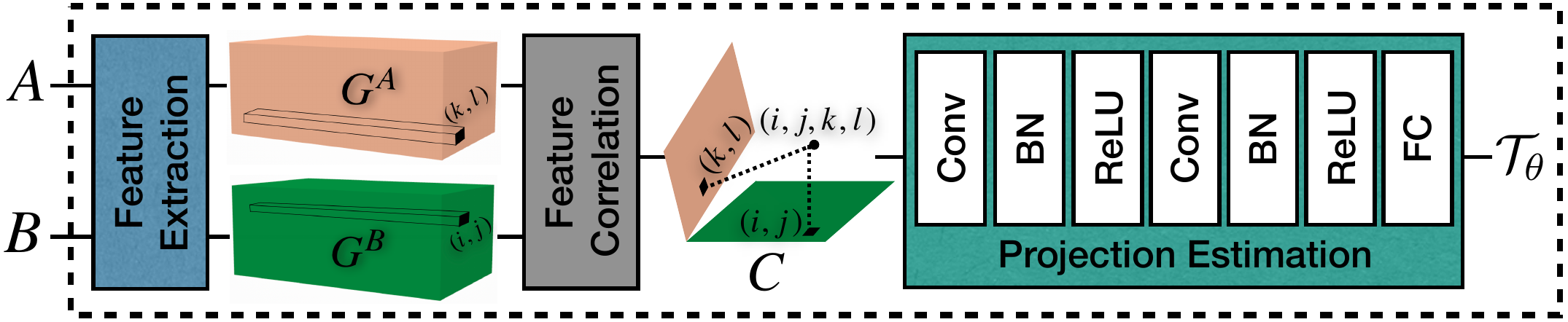}
	\caption{Geometric warping Network; it runs parallel to the texture transfer network.}
	\label{fig:composition-align}
\end{figure*}

\section{Geometric Style Transfer}\label{sec:com-align}

Geometric style transfer warps the content image onto 
the style image before texture is transferred. 
GST comprises of two parts, 
as illustrated in Figure~\ref{fig:composition-align}.
First, feature correlation measures the degree 
to which geometric features (see Section~\ref{sec:fea-extrac}) 
in the content and style image correlate. 
Second, the correlations are input to a trained network that provides a first approximation 
to a geometric mapping $\mathcal{M}: \Re^2 \mapsto \Re^2$; 
this mapping is governed by geometric style. We use parametric spatial mappings; 
we've found either affine transforms or quadratic thin-plate to be sufficient 
but there is no reason in principle not to use other mappings.
Each part is described in detail below.

\subsection{Feature Correlation} 
Feature correlation is between the geometric feature sets, 
$G^c$ and $G^s$ defined in Section~\ref{sec:fea-extrac}. These are each arrays of size $W_4 \times H_4$ and contain vector elements of length $N_4$, Elements in these sets are indexed by their location in the sample array. Then correlation function is a four-dimensional function:
\begin{equation}
C_{i,j,k,l}=  \hat{f}^c_{i,j}\odot \hat{f}^s_{k,l}, \label{eqn:corr}
\end{equation}
with $\odot$ being the inner product between vectors,
and $\hat{.}$ indicating $L_2$ normalization.
$C$ is then postprocessed to zero out negative values.
A visualization of this procedure is illustrated in Figure~\ref{fig:composition-align}.

\subsection{Learning a Spatial Mapping}\label{ProjectionEstimation} 

To transfer geometric style,
we require a spatial transform $\mathcal{T}: \Re^2 \mapsto \Re^2$
to warp the content image to the style image.
We assume a parametric transform,
and therefore use a regression CNN to determine parameter values, $\Theta$.
At this stage we are computing a first approximation to
the spatial transform and have found an affine transform;
in the next step will will upgrade to a thin-plate spline.
Whether we use these or some other transform,
we wish to compute the mapping parameters using the correlation matrix, $C$,
so that $\Theta = \mu(C)$.
To do this, we use a regression CNN to learn the mapping
$\mu: \Re^{W \times H \times W \times H} \mapsto \Re^{p}$
where $p$ is the number of parameters,
and $W,H$ control the grid of sample sites.

The regression CNN is trained by iterating over many identical trials. 
At the $i^{th}$ trial we randomly sample transform parameters $\Theta_i$, 
sampling details are given below. 
The parameters specify a spatial transform we will call $\mathcal{T}_i$. 
Each ground truth transform is used to warp training images $A$ to make $B = \mathcal{T}_i(A)$, 
and also to move sample locations for geometric features, $(x,y)_{wh}$,  
to get sample locations in the warped image at $\mathcal{T}_i (x,y)_{wh}$.
The image $B$ is then processed using Gatys~\etal ~\cite{gatys2016image}
to create a artistic-texture-augmented copy.
We can now  use  the original image $A$ and the 
warped and artistic-texture augmented
image $B$ to compute a correlation matrix as described by Equation~\ref{eqn:corr}; 
we will call this correlation $C_i$. 
This process explicitly connects the known parameters $\Theta_i$ 
to a known correlation matrix $C_i$. 
The quality of the (current) mapping $\mu$ is measured 
using the L2-norm between the sample locations 
mapped under the known transform, $\mathcal{T}_i$, 
and the transform constructed from the parameters $\mu(C_i)$, 
which we write at $\mathcal{T}_{\mu|C_i}$.
Thus, the regression net has the loss function:
\begin{equation}\label{eq:loss1}
    \mathcal{L} =  \sum_i || \mathcal{T}_{\mu|C_i}(x_{jk}), \mathcal{T}_i(x_{jk}) ||_2.
\end{equation}
Once trained, the network will compute spatial transform parameters, given a correlation matrix $C_{ijkl}$.

The learning process above will work in principle for any transform.
Even so, finding the optimal thin-plate spline (TPS) 
transformation~\cite{bookstein1989principal} is not easy. 
We have found it useful in practice to learn such higher-order mappings  by first estimating an affine mapping using the above, 
and using this affine mapping to warp the image $A$ before then 
warping it a second time using the higher-order mapping that is being learned. Note that this requires two copies of the geometric warping network: the first outputs the 6 parameters of an affine mapping, the second outputs the 18 parameters of a quadratic thin-plate spline. The two networks are distinct.

\smallskip
\noindent{\bf Comments on Geometric Warping:} The examples in this paper all use either affine or TPS warping. However, there is nothing about the architecture that limits it to that pair of warping families. We could have used bicubic warps, or a homography, for example.  The geometric warping network could sit in parallel to many existing texture transfer architectures. The reader is free to implement our network alongside their own, and to explore different families of geometric warp.

\begin{figure}[bp]
	\centering
	\includegraphics[width=0.8\linewidth]{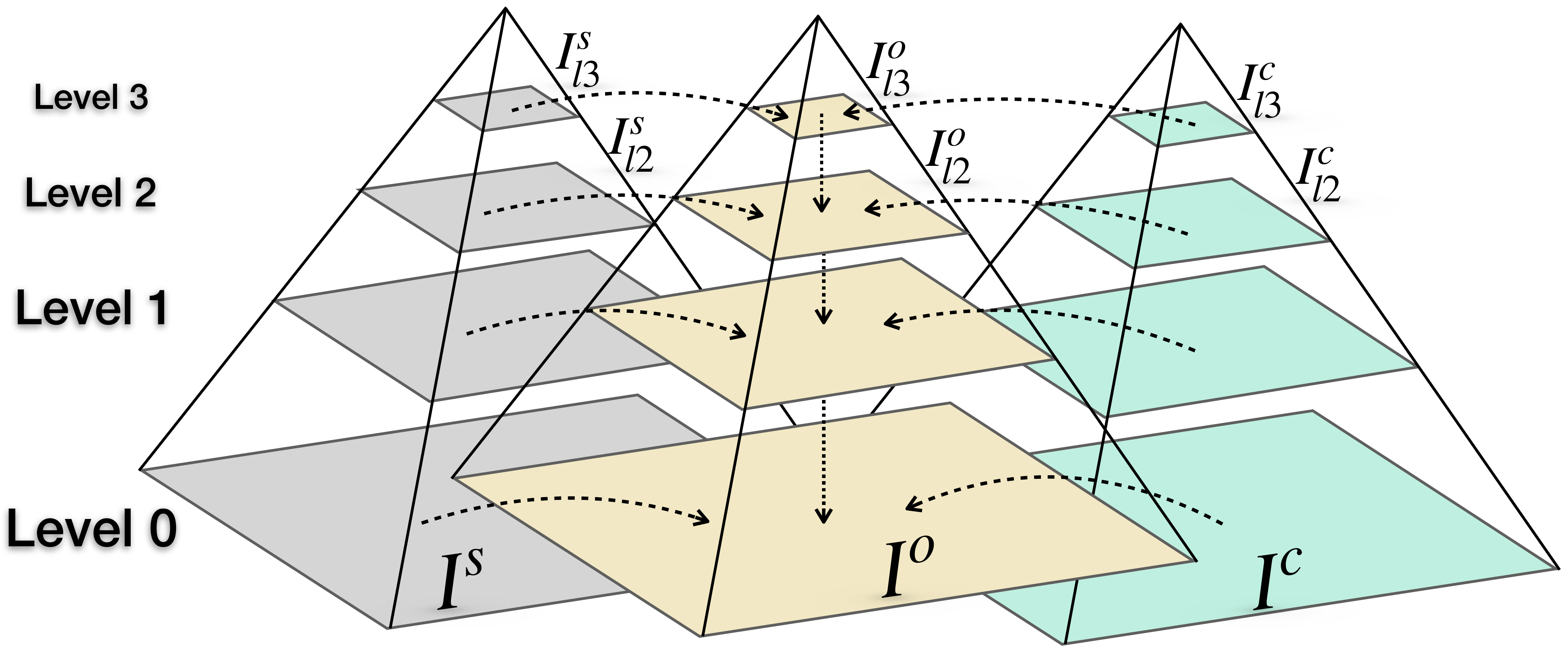}
	\caption{Multi-scale strategy used in style texture transfer.}
	\label{fig:multi-scale}
\end{figure}

%%%%%%%%%%%%%%%%%%%%%%%%%%%%%%%%%%%%%%%%%%%%%%%%%%%%%%%%%%%%
\section{Texture Transfer}\label{sec:style-trans}

Texture style transfer is used to approximate denotational style to input images. In our case these will be a content image that has been warped by the geometry style network of the previous section.
Let $I^o$ be the generated stylized result,
and $I^s$ and $I^c$ are the style and content image respectively;
the content image has been warped to match geometric style.
As stated in Section~\ref{sec:fea-extrac},
the style of an image is represented by the texture feature $D$.
The texture style reconstruction loss is a weighted sum of L2-norms:
\begin{equation}\label{eq:lstyle}
\mathcal{L}_{texture}(I^s,I^o)=\frac{1}{2}\sum_{l}\omega_l\| D_{l}^s-D_{l}^o\|_2,
\end{equation}
where $l\in\{1, 2, 3, 4, 5\}$, stands for the selected layers,
and $\omega_l$ is the weighting factor for each layer.

The content of an image is represented by 
the content feature $F$ (see Section~\ref{sec:fea-extrac}),
and the content reconstruction loss is the L2-norm of two features:
\begin{equation}
\mathcal{L}_{content}(I^c,I^o)=\frac{1}{2}\| F^c-F^o\|_2.
\end{equation}

The loss function we minimise is:
\begin{equation}\label{eq:total-loss}
\mathcal{L}_{total}=\alpha\mathcal{L}_{texture}(I^s,I^o)+\beta\mathcal{L}_{content}(I^c,I^o),
\end{equation}
where $\alpha$ and $\beta$ are the weighting factors.

\begin{figure*}[tbp]
	\centering
	\includegraphics[width=\linewidth]{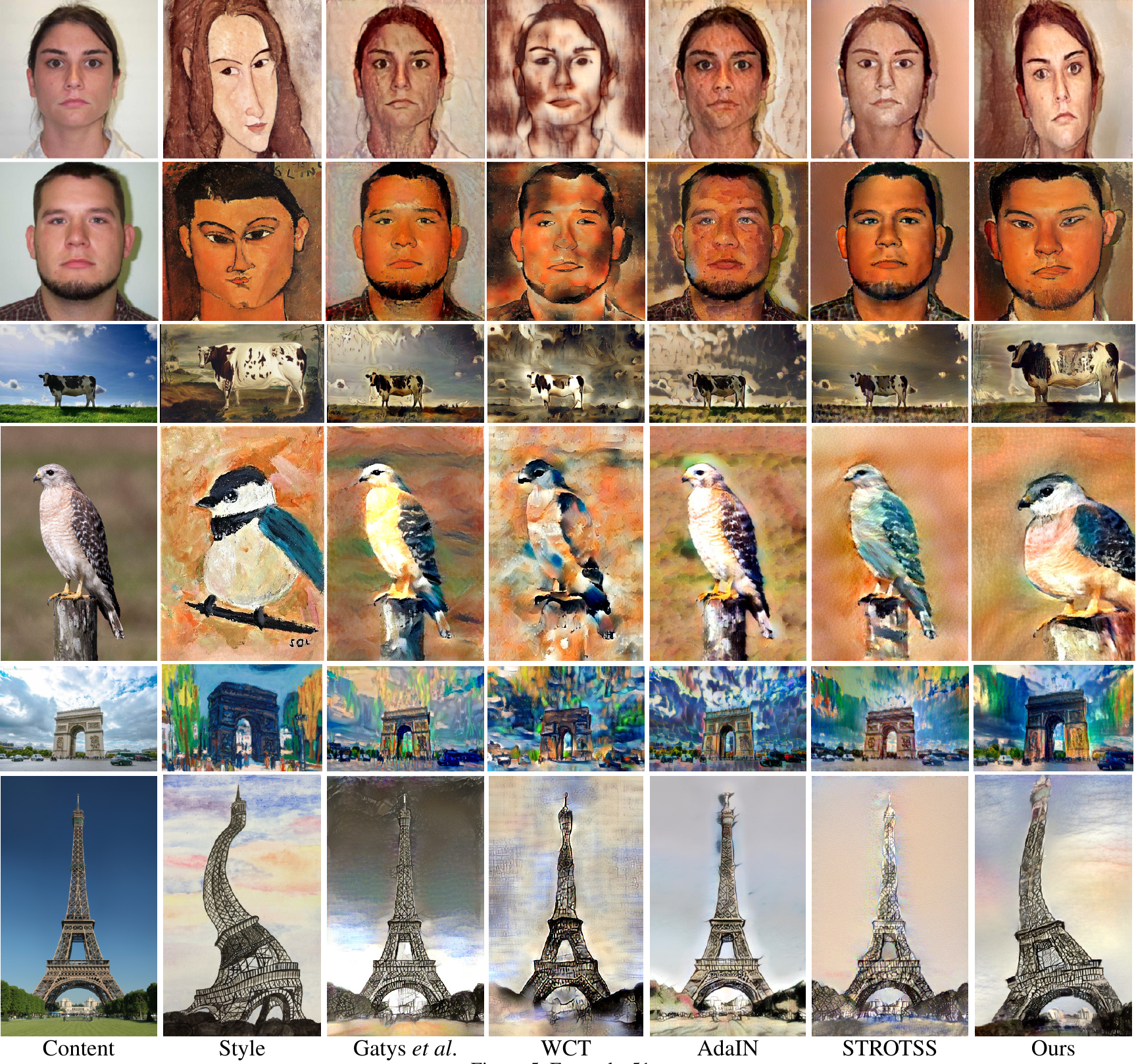}
	\caption{Comparison to related methods. Gatys~\etal~\cite{gatys2016image} and STROTSS~\cite{kolkin2019style} are image optimization based,
    AdaIN~\cite{huang2017arbitrary} and WCT~\cite{li2017universal} are feature transformation based.}
	\label{fig:compare1}
\end{figure*}

\smallskip
\noindent{\bf Multi-Scale Strategy:}\label{sec:multi-scale}
After geometric style transfer (Section~\ref{sec:com-align}),
some parts of the image will be enlarged in some cases (see Figure~\ref{fig:ablation2}),
and the resolution will decrease accordingly.
Furthermore, as pointed out in previous work~\cite{luo2016understanding},
the effective receptive field of network neurons is fixed and relatively small, 
which limits the scale of synthesized features.
In order to avoid the stylized effect being affected by the decreased image resolution we adopt a multi-scale strategy to do style transfer, inspired by prior art~\cite{heeger1995pyramid,Han:2008,Snelgrove:2017,gatys2017controlling}
that shows good texture synthesis with a bank of multi-scale filters.
Specifically, 
as shown in Figure~\ref{fig:multi-scale},
we first downsample $I^s$ and $I^c$ by feeding them into a 
Gaussian pyramid~\cite{adelson1984pyramid} (here we use $3$ levels),
and use $I^s_{l3}$ and $I^c_{l3}$ to perform stylization and get $I^o_{l3}$.
Then we upsample $I^o_{l3}$ as a initialization,
utilize $I^s_{l2}$ and $I^c_{l2}$ to generate $I^o_{l2}$, and so on. 
In this way, during the stylization,  
the optimization will simultaneously match features at every pyramid layer,
which guarantees the generation of high-resolution outputs 
(we further discuss this in Section~\ref{sec:ablation}).

\begin{figure*}[htbp]
	\centering
	\includegraphics[width=\linewidth]{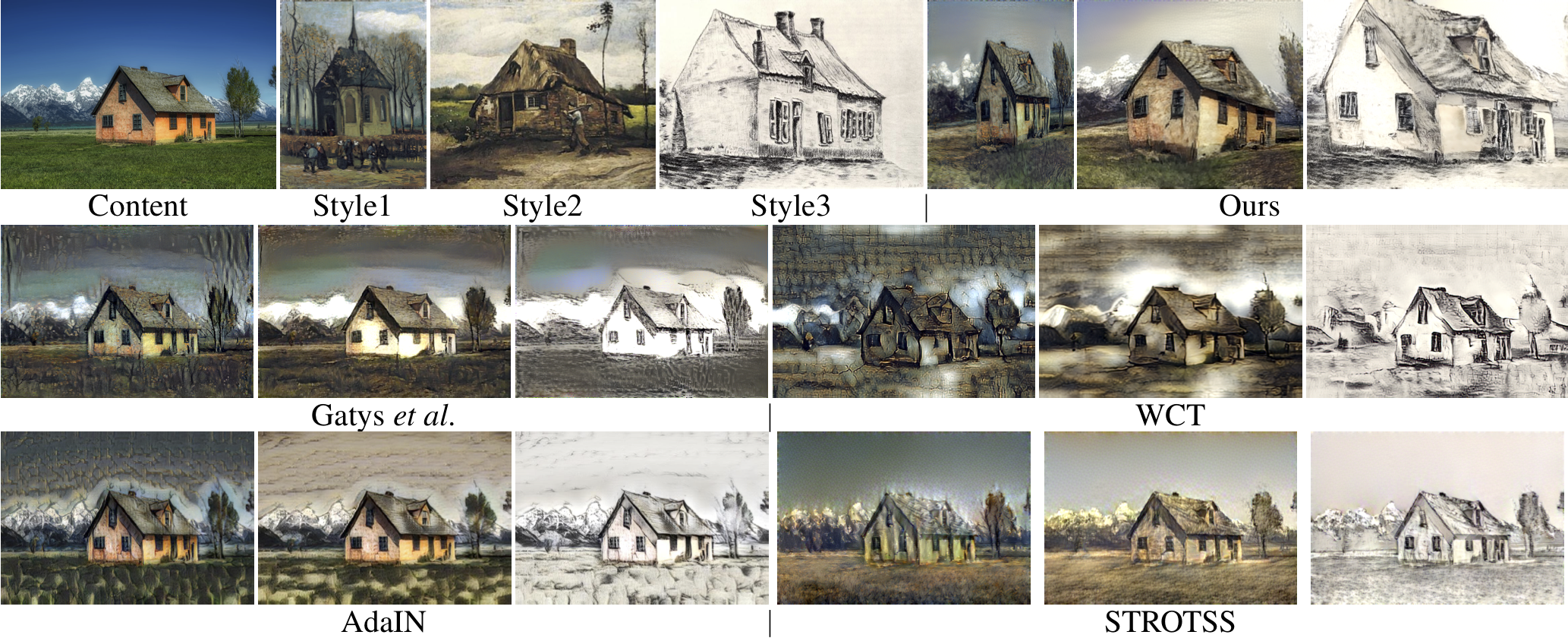}
	\caption{Comparison on one content with multiple style images, the geometric styles will change with different style images.}
	\label{fig:compare2}
\end{figure*}

%%%%%%%%%%%%%%%%%%%%%%%%%%%%%%%%%%%%%%%%%%%%%%%%%%%%%%%%%%%%%%%%%%%%

\section{Experimental Results}
\label{sec:experimental-results}

Our experimental results come in three parts: (i) qualitative results designed to show the reader the difference that geometry transfer makes to style transfer; (ii)  quantitative experiments that place a subjective measure over the degree of difference; (iii) ablation studies to show the difference between affine and TPS geometric style transfer and of multi-scale texture transfer.
Experiments and results are explained below, before which we provide implementation details.

Our implementation uses PyTorch~\cite{paszke2017automatic}.
We use the pre-trained VGG-19 exactly as described in~\cite{simonyan2014very}. 
The geometric warping network is trained on the Microsoft COCO dataset~\cite{lin2014microsoft}.
We resize each of the training images to $240\times240$ 
and train the network with a batch size of $8$. 
We use Adam~\cite{kingma2014adam} with a learning rate of $1\times10^{-3}$;
training takes roughly 8 hours on a single GTX 1080Ti GPU;
Sample points $x_{jk}$ in Equation~\ref{eq:loss1} are from a $20\times20$ uniform grid.
For multi-scale texture transfer (Section~\ref{sec:style-trans}), 
we weight each layer equally in Equation~\ref{eq:lstyle} ($\omega_l$=1/5),
the ratio $\alpha/\beta$ in Equation~\ref{eq:total-loss} is $5\times10^{-3}$. Run-times are comparable with other image optimization methods, generating a $512\times512$ image takes around 50 seconds.

\subsection{Qualitative Comparisons}

We qualitatively compare our method with some closely related NST approaches. We could not compare with NST methods that deal with geometric transfer~\cite{Yaniv2019,yuphoto,li2018carigan,cao2018carigans,shi2019warpgan, Yang_textNST_2019_ICCV} because (a) they each deal with one object class only (faces or text) and (b) they tend not to conform to the `content/style' input paradigm. Instead, we compare with well known  methods that like us input a single content photograph and a single style image, and which are intended to be general purpose. We compare to Gatys~\etal~\cite{gatys2016image} and STROTSS~\cite{kolkin2019style}
which are image optimization methods; and to 
AdaIN~\cite{huang2017arbitrary} and WCT~\cite{li2017universal}
which are feature transformation methods. Results are shown in Figures~\ref{fig:compare1} and~\ref{fig:compare2}.

From the comparison results,
we should first notice that for other methods,
the output sizes are the same as that of content images,
while the size of our results match that of style images.
Second, from the view of artistic effects,
all the results keep texture features and color distributions well.
The most striking difference between our output and that of all other algorithms is that only ours changes  shapes within the content image. Our output portraits lengthen the face when necessary, skew facial features when that is part of the style, makes livestock larger, fattens birds, and bends towers and houses.

%-------------------------------------------------------------
\subsection{Quantitative Comparisons}

Here we present quantitative results relating to the quality of output, the impact of geometric style, and computational efficiency.

\smallskip
\noindent{\bf Output Quality:} 
There is no objective measure by which to assess the quality of outputs, therefore we followed others by conducting a questionnaire investigation to survey the preferences of different approaches.
Every questionnaire included 10 pairs of content-style pairs, 
and participants are asked to vote for their favorite results.
We collected questionnaires from $50$ respondents 
and computed the percentage of every method
with regard to the preferences. Results are shown in Figure~\ref{fig:user-preference}.
Our results are preferred more than any other, at 47\% we are more than twice as likely as the next most popular, of 18\%. However, popularity makes not statement about success in reaching the target style, our next experiment was designed to address this.

\definecolor{a}{RGB}{0, 162, 255}
\definecolor{b}{RGB}{97, 216, 54}
\definecolor{c}{RGB}{248, 186, 0}
\definecolor{d}{RGB}{255, 38, 0}
\definecolor{e}{RGB}{194, 72, 133}
\definecolor{f}{RGB}{153, 150, 146}
\newcommand\filledcirca{{\color{a}\bullet}\mathllap{\color{a}\circ}}
\newcommand\crule[1][black]{\textcolor{#1}{\rule{0.3cm}{0.3cm}}}

\begin{figure}[htbp]
\hspace{3mm}
\begin{minipage}{0.4\linewidth}% adapt widths of minipages to your needs
	\includegraphics[width=\linewidth]{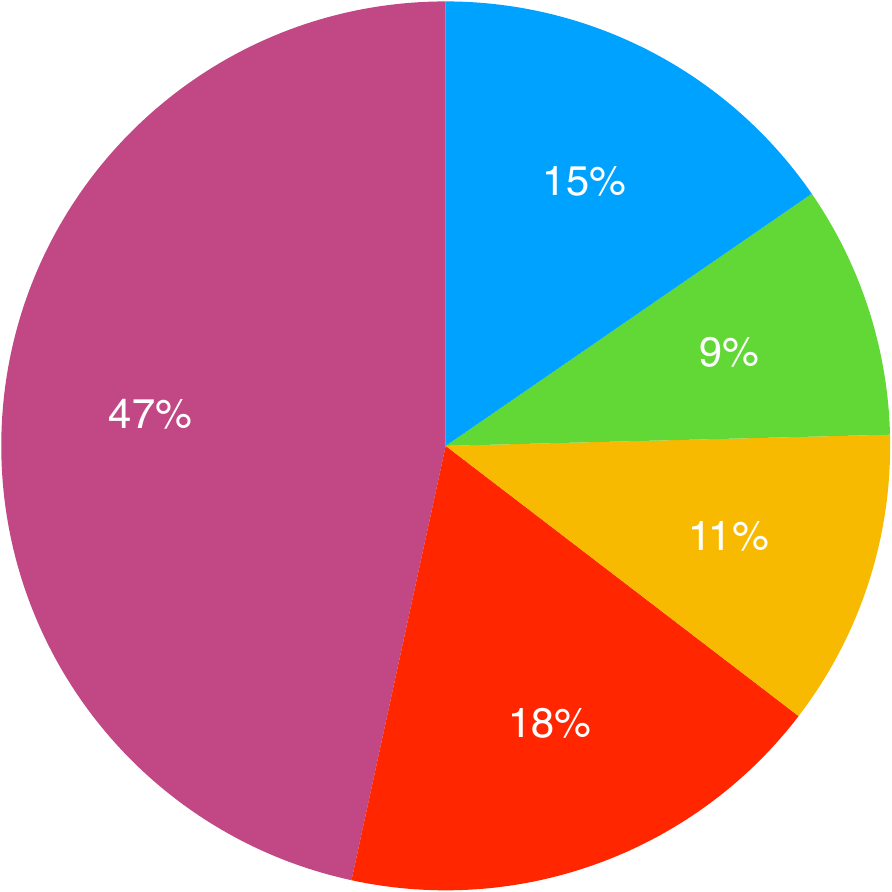}
\end{minipage}%
\hspace{3mm}
\begin{minipage}{0.4\linewidth}\raggedright
	\crule[a] Gatys~\etal~\cite{gatys2016image}\\ \vspace{2.5mm}
	\crule[b] AdaIN~\cite{huang2017arbitrary}\\ \vspace{2.5mm}
    \crule[c] WCT~\cite{li2017diversified}\\ \vspace{2.5mm}
	\crule[d] STROTSS~\cite{kolkin2019style}\\ \vspace{2.5mm}
	\crule[e] Ours
\end{minipage}
\caption{Participant preferences -- our output is preferred more than twice as much as the nearest alternative.}
\label{fig:user-preference}
\end{figure}

\smallskip

\begin{figure}[htbp]
	\centering
	\includegraphics[width=0.9\linewidth]{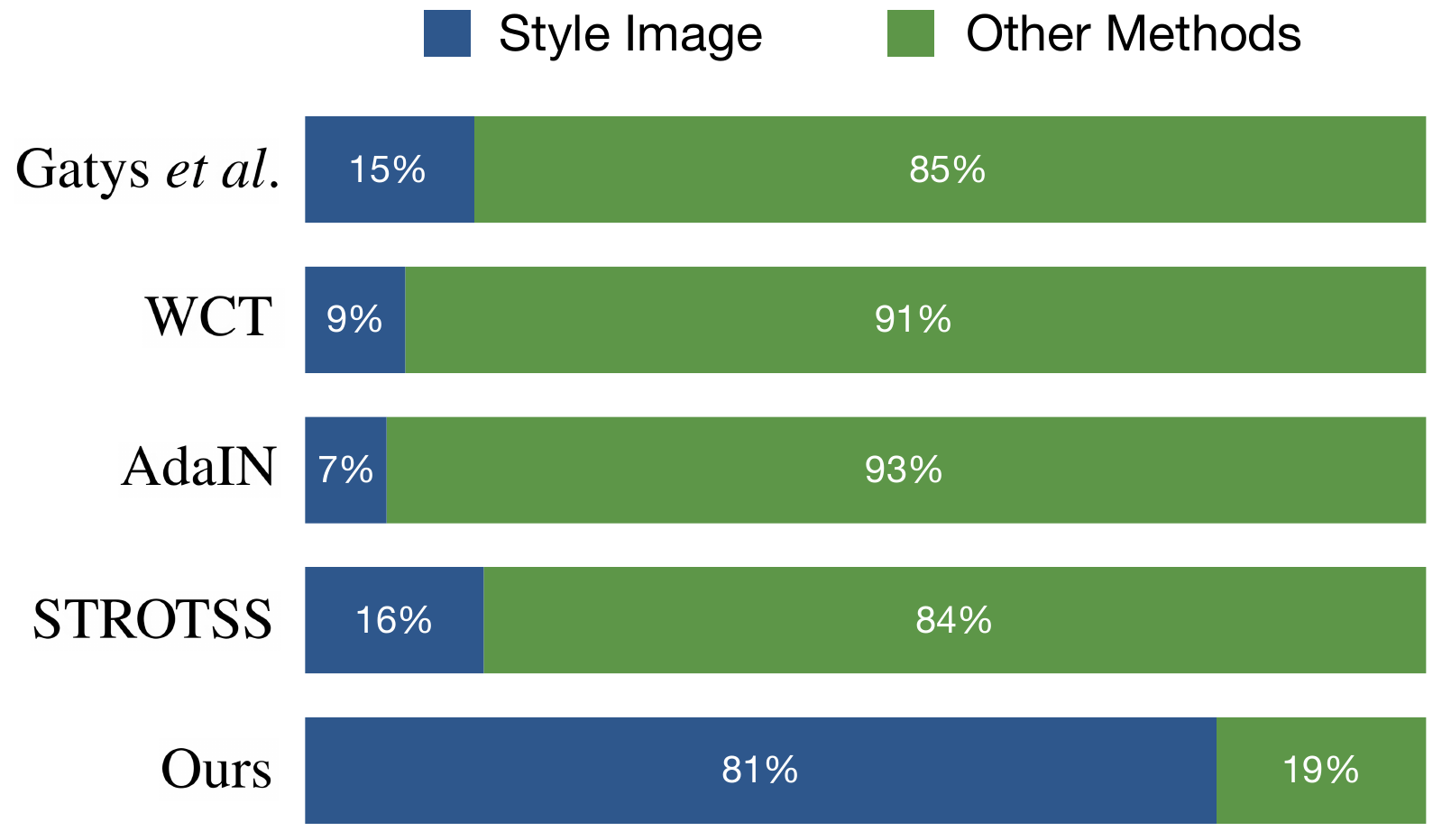}
	\caption{Results from our similarity experiment: output from four NST algorithms without GST are more likely to be assessed as more similar to each other than to the style image. Our GST output is judged as being closer to the style image.}
	\label{fig:method-similarity}
\end{figure}

\noindent{\bf  Output Similarity:} 
The preferences of participants measure popularity but say nothing about the closeness of output to the target style. Asking questions about the similarity of output to the target style helps give us a handle on success in that regard.
We showed 50 participants 3 images; one of the 3 was the style image,  the other two images came from one of five algorithms, Gatys~\etal~\cite{gatys2016image},  STROTSS~\cite{kolkin2019style}
AdaIN~\cite{huang2017arbitrary}, WCT~\cite{li2017universal}, and ours. The three images, call them A,B,C were shown as three side-by-side pairs (A,B), (A,C), (B,C), each pair on a separate row. The choice of algorithm, the location of the images pairs and the ordering of a pair were all subject to randomization. We give each participant this simple instruction: {\em check the pair of images you think are most similar}. No other information  was given to the participant nor did we ask any participant to explain their preference.

We recorded all preferences, and the number of times an image from an algorithm was picked. We collated this data into (a) the fractional number of times an output image from an algorithm was regarded as more similar to the style image, and (b) the fractional number of times the output images were regarded as more similar to each other.  Figure~\ref{fig:method-similarity} shows results in percentage form. Participants judged our output to be closer to the style target about 81\% of the time, compared to at most 16\% for any other. Furthermore, the other algorithms are more likely to be judged as more similar to each other than the style image. The standard deviation is about 4\%. 
% See supplementary material for raw results.

The fact that the participants judge other algorithms' outputs to be closer to each other than to the style image is important: the algorithms produce different textural output, so whatever criteria the participants used to assess similarity, it must have a stronger influence than texture. The fact our output was very much more likely to be chosen as most similar to the style target, strongly suggests that geometric style transfer explains the result.

\begin{figure}[tbp]
	\centering
	\includegraphics[width=1.0\linewidth]{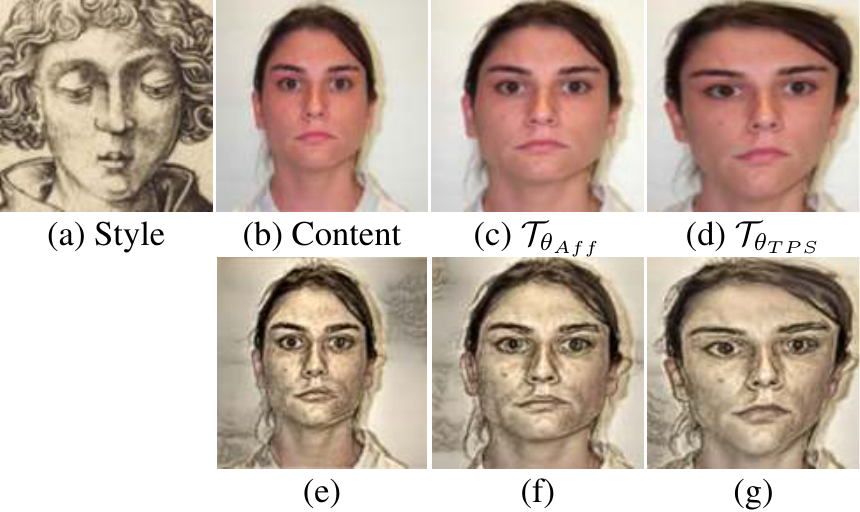}
	\caption{Effects of geometric warping.
		(a) style image, (b) content image, (c) warping only using 
		an affine transformation $\mathcal{T}_{\theta_{Aff}}$,
		(d) warping with TPS transformations $\mathcal{T}_{\theta_{TPS}}$.
		(e) (f) and (g) are the texture transfer results of (b) (c) and (d) respectively.
		Affine transformation provides a rough warping but 
		keeps some invariants (\eg parallel lines, ratio of areas), 
		 TPS refines the transformation to a better match.}
	\label{fig:ablation1}
\end{figure}

\subsection{Discussions and Ablation Studies}\label{sec:ablation}

In this part we discuss two factors that highly influence the results:
Geometric warping (Section~\ref{ProjectionEstimation})
and Multi-scale strategy (Section~\ref{sec:multi-scale}),
as well as the limitations of our method.

\smallskip
\noindent{\bf Geometric Warping.}
As stated in Section~\ref{ProjectionEstimation},
the final estimated transformation either an affine transformation $\mathcal{T}_{\theta_{Aff}}$
or  a thin-plate spline (TPS) transformation $\mathcal{T}_{\theta_{TPS}}$. 
Figure~\ref{fig:ablation1} illustrates the effectiveness of two transformations.
We can see that affine transformation  moves and scales 
the source image to roughly match the target,
but the invariant properties of the affine transformation, 
(\eg parallel lines, ratio of areas) prevent it from reaching a closer geometric mapping. The TPS refines the transformation to generate a better warping. As noted previously, there is nothing to prevent our approach reaching higher-order transforms, but we have not found it necessary.

\begin{figure}[ht]
	\centering
    \begin{overpic}[width=\linewidth]{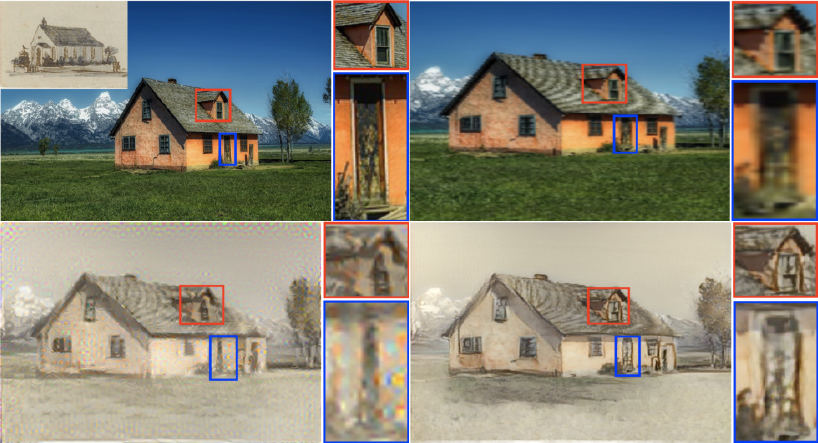}
		\put(1,28){(a)}
		\put(51,28){(b)}
		\put(1,1){(c)}
		\put(51,1){(d)}
	\end{overpic}
	\caption{Style transfer with multi-scale strategy. 
	(a) Content and style (upper left) images.
	(b) Output after the geometric warping. (c) Output without multi-scale synthesis.
	(d) Output with multi-scale strategy. The colored boxes show the magnified details.
	(b) shows that the resolution will decrease after the geometric warping,
	and details in (c) will lose accordingly.
	Multi-scale strategy (Section~\ref{sec:multi-scale}) will ensure the 
	generation of high-resolution results without losing too much detail.}
	\label{fig:ablation2}
\end{figure}

\smallskip
\noindent{\bf Multi-Scale Strategy. }
This strategy was applied to improve the quality 
of the stylized results (Section~\ref{sec:multi-scale}). This is important to us, because sometimes the geometric alignment 
manipulation will enlarge parts of the image, which reduced image sharpness accordingly. This is seen in  Figure~\ref{fig:ablation2}:
the output is not sufficiently sharp and some details are missing unless the multi-scale approach is used.

\begin{figure}[htbp]
    \centering
    \includegraphics[width=\linewidth]{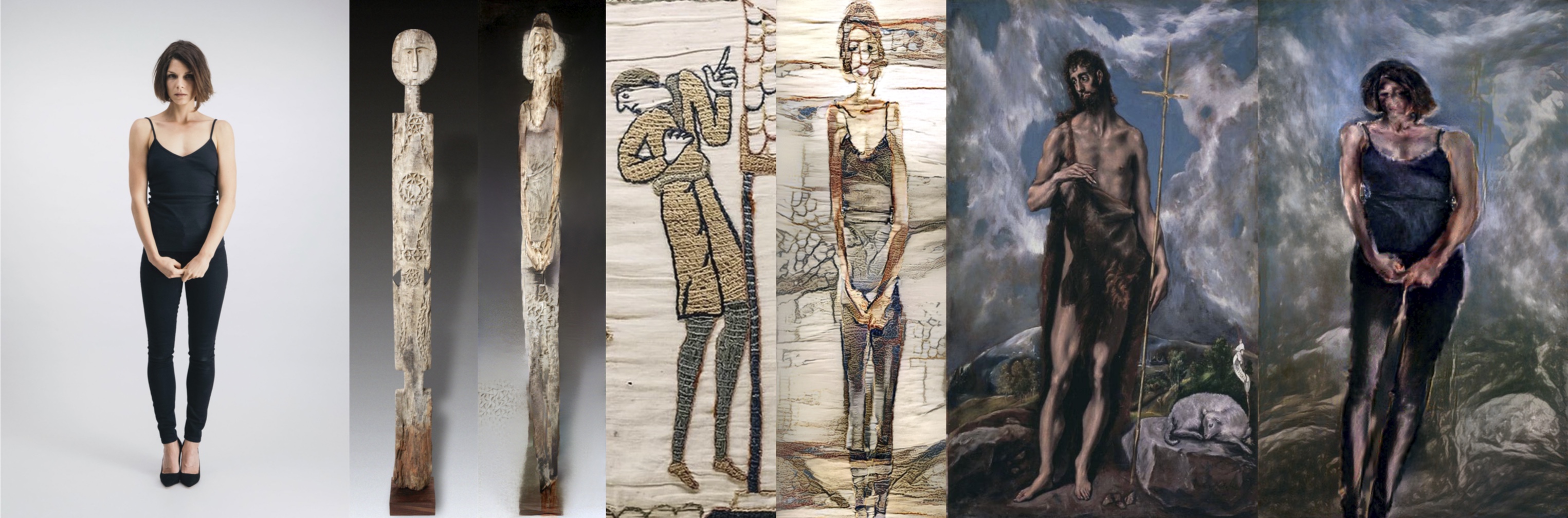}
    \caption{Some interesting cases, left to right, content image then 3 styles: African, Bayeux tapestry detail, el Grecco. In each case, the style image is on the left, output on the right.}
\label{fig:limits}
\end{figure}

\noindent{\bf Limitations and Interesting Cases:}  Limiting assumptions are: (i) the content image and geometric style images  share same semantic content and each show one major object; (ii) the geometric style  can be matched using a continuous warping functions across the whole image. Assumption (i) is required for feature matching. Assumption (ii) confines the geometric styles we can reach beyond changing to higher-order maps. Some styles such as Cubism require piece-wise spatial mappings; other artists often change use of orthogonal projection, or use many vanishing points; pose can be un-natural, as in Egyptian art.

Limiting case (i) is less limiting than it sounds. First of all, the method itself is agnostic with respect to image content, but best results are to be had with similar semantic content. To see why this is the case recall the fact that human artists alter geometry for emphasis and in non-arbitrary ways --  Stubbs  exaggerated bull-like characteristics to depict bigger, stronger animals. Similarly, faces are altered to bring out some desired latent character, such as femininity or masculinity. This means that the manner of warp is class-conditional: human artists do not usually try to distort horses towards houses. It is, therefore, not at all unreasonable for the geometric style picture to contain an exemplar related to the object class in the content image.

Interesting examples arise even when these conditions are adhered to, as  Figure~\ref{fig:limits} shows. A global spatial transform means that local pose changes \etc are not well modeled, noticeable in the Bayeux tapestry detail. El Grecco, known for stylized elongation of bodies, also has a pose change but our output suffers less, probably because the change of pose has little impact on the overall profile. Similar remarks apply to the African sculpture.

Important detail is not always transferred well. Facial features are lost in all cases, the Bayeux tapestry detail is not a convincing tapestry; and the African sculpture appears weathered.
Where the content image is blank, our algorithm copies texture more-or-less directly from the texture image. Some other algorithms also do this, see Figure~\ref{fig:compare1} for examples.

\begin{figure}[tbp]
    \centering
    \includegraphics[width=\linewidth]{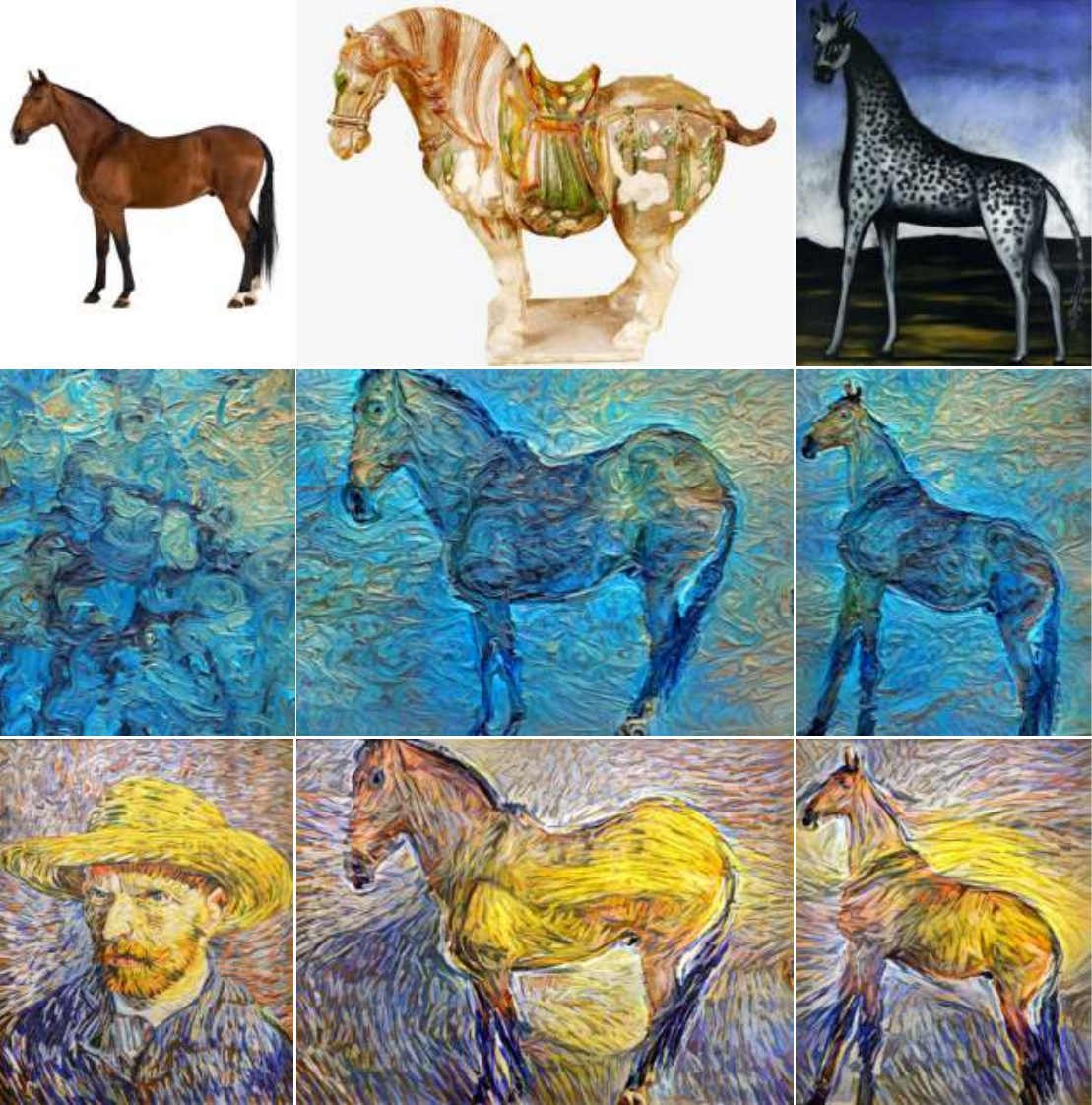}
    \caption{Style transfer using three input images. Content image, top left; geometry style images top-middle and top-right; texture style images left-middle and left-bottom. Output images in the corresponding $2 \times 2$ array. }
    \label{fig:3imagin}
\end{figure}

Finally, we do not have to use a single style image but can instead use two style images: one for geometric style the other for texture style, Figure~\ref{fig:3imagin} shows examples.  The figure has one content image, in the top-left. Geometric style images are placed top-middle and top-right, with texture style images on the left column. These reference images form a $2 \times 2$ array of corresponding output images. 

Using three images extends the current paradigm in a novel and useful way. This adds versatility to the system because different pictures can be used to specify different components of style. For example, as in Figure~\ref{fig:3imagin}, pure texture can be used to specify the texture style, and statues can be used to specify geometry style. This would not be possible using a single image to specify style. The principle might be extended in the future so that different elements of output style (\eg texture, geometry, composition) are characterized by different example images.

\section{Conclusions}

Our paper provides a novel method for image stylization: geometric style transfer. 
We provide a network to compute geometric style in a general setting, and use multi-scale texture transfer to maintain image quality throughout the transfer. Experimental results illustrate the qualitative expressiveness of our stylized results and greater quantitative similarity to target styles than other algorithms' outputs. These results are consistent with the Art History literature,  where projection style has been used to characterise human art~\cite{willats1997art}.

Our algorithm does have limits that provides plenty of future work. 
The content of the style image must be similar enough to the content
of the content image for high-level features to be matched. 
This is more general than the requirement of strong models of faces or text~\cite{yuphoto,li2018carigan,cao2018carigans,shi2019warpgan,Yaniv2019,Yang_textNST_2019_ICCV}, but it is nonetheless a restriction. 
Our algorithm is global, whereas many styles will be local. 
This is known to be the case for texture~\cite{Kotovenko_2019_ICCV} and facial features~\etal~\cite{Yaniv2019}. Some styles, such as Cubism, are beyond out scope --  but this is true of every other NST algorithm we know of.  If NST is to progress to styles of that kind, then algorithms that include geometric style transfer are inevitable.

{\small
\bibliographystyle{ieee_fullname}
\bibliography{egbib}

\begin{thebibliography}{10}\itemsep=-1pt

\bibitem{adelson1984pyramid}
Edward~H Adelson, Charles~H Anderson, James~R Bergen, Peter~J Burt, and Joan~M
  Ogden.
\newblock Pyramid methods in image processing.
\newblock {\em RCA engineer}, 29(6):33--41, 1984.

\bibitem{arnheim1954art}
Rudolf Arnheim.
\newblock {\em Art and visual perception: A psychology of the creative eye}.
\newblock Univ of California Press, 1954.

\bibitem{bookstein1989principal}
Fred~L. Bookstein.
\newblock Principal warps: Thin-plate splines and the decomposition of
  deformations.
\newblock {\em IEEE Transactions on Pattern Analysis and Machine Intelligence},
  11(6):567--585, 1989.

\bibitem{cao2018carigans}
Kaidi Cao, Jing Liao, and Lu Yuan.
\newblock Carigans: Unpaired photo-to-caricature translation.
\newblock {\em ACM Transactions on Graphics (Proc. of Siggraph Asia 2018)},
  2018.

\bibitem{champ2016semantic}
Alex~J. Champandard.
\newblock Semantic style transfer and turning two-bit doodles into fine
  artworks.
\newblock {\em \tt arXiv:1603.01768 [cs.CV]}, 2016.

\bibitem{chen2017coherent}
Dongdong Chen, Jing Liao, Lu Yuan, Nenghai Yu, and Gang Hua.
\newblock Coherent online video style transfer.
\newblock In {\em IEEE International Conference on Computer Vision (ICCV)},
  pages 1105--1114, 2017.

\bibitem{chen2017stylebank}
Dongdong Chen, Lu Yuan, Jing Liao, Nenghai Yu, and Gang Hua.
\newblock {StyleBank:} an explicit representation for neural image style
  transfer.
\newblock In {\em IEEE Conference on Computer Vision and Pattern Recognition
  (CVPR)}, pages 1897--1906, 2017.

\bibitem{chen2018stereoscopic}
Dongdong Chen, Lu Yuan, Jing Liao, Nenghai Yu, and Gang Hua.
\newblock Stereoscopic neural style transfer.
\newblock In {\em IEEE Conference on Computer Vision and Pattern Recognition
  (CVPR)}, pages 6654--6663, 2018.

\bibitem{chen2016towards}
Yi-Lei Chen and Chiou-Ting Hsu.
\newblock Towards deep style transfer: A content-aware perspective.
\newblock In {\em British Machine Vision Conference (BMVC)}, 2016.

\bibitem{collomosse2003cubist}
John~P Collomosse and Peter~M Hall.
\newblock Cubist style rendering from photographs.
\newblock {\em IEEE Transactions on Visualization and Computer Graphics},
  9(4):443--453, 2003.

\bibitem{decarlo2002stylization}
Doug DeCarlo and Anthony Santella.
\newblock Stylization and abstraction of photographs.
\newblock {\em ACM Transactions on Graphics (ToG)}, 21(3):769--776, 2002.

\bibitem{deng2009imagenet}
Jia Deng, Wei Dong, Richard Socher, Li-Jia Li, Kai Li, and Li Fei-Fei.
\newblock Imagenet: A large-scale hierarchical image database.
\newblock In {\em IEEE Conference on Computer Vision and Pattern Recognition
  (CVPR)}, pages 248--255, 2009.

\bibitem{Gatys2015c}
L.~A. Gatys, A.~S. Ecker, and M. Bethge.
\newblock A neural algorithm of artistic style.
\newblock {\em \tt arXiv:1508.06576[cs.CV]}, 2015.

\bibitem{gatys2016image}
Leon~A Gatys, Alexander~S Ecker, and Matthias Bethge.
\newblock Image style transfer using convolutional neural networks.
\newblock In {\em IEEE Conference on Computer Vision and Pattern Recognition
  (CVPR)}, pages 2414--2423, 2016.

\bibitem{gatys2017controlling}
Leon~A Gatys, Alexander~S Ecker, Matthias Bethge, Aaron Hertzmann, and Eli
  Shechtman.
\newblock Controlling perceptual factors in neural style transfer.
\newblock In {\em IEEE Conference on Computer Vision and Pattern Recognition
  (CVPR)}, pages 3985--3993, 2017.

\bibitem{gupta2017characterizing}
A. Gupta, Justin Johnson, Alexandre Alahi, and Li Fei-Fei.
\newblock Characterizing and improving stability in neural style transfer.
\newblock In {\em IEEE International Conference on Computer Vision (ICCV)},
  pages 4067--4076, 2017.

\bibitem{haeberli1990paint}
Paul Haeberli.
\newblock Paint by numbers: Abstract image representations.
\newblock In {\em Proceedings of the 17th Annual Conference on Computer
  Graphics and Interactive Techniques (SIGGRAPH)}, pages 207--214. ACM, 1990.

\bibitem{hall1998example}
P.M. Hall.
\newblock Painting by example.
\newblock In {\em Eurographics UK}, pages 159--167, 1998.

\bibitem{hall2007rtcams}
Peter~M Hall, John~P Collomosse, Yi-Zhe Song, Peiyi Shen, and Chuan Li.
\newblock Rtcams: A new perspective on nonphotorealistic rendering from
  photographs.
\newblock {\em IEEE Transactions on Visualization and Computer Graphics},
  13(5):966--979, 2007.

\bibitem{Han:2008}
Charles Han, Eric Risser, Ravi Ramamoorthi, and Eitan Grinspun.
\newblock Multiscale texture synthesis.
\newblock {\em ACM Transactions on Graphics (ToG)}, 27(3):51:1--51:8, 2008.

\bibitem{heeger1995pyramid}
David~J Heeger and James~R Bergen.
\newblock Pyramid-based texture analysis/synthesis.
\newblock In {\em Proceedings of the 22nd Annual Conference on Computer
  Graphics and Interactive Techniques (SIGGRAPH)}, pages 229--238. ACM, 1995.

\bibitem{hertzmann2003survey}
Aaron Hertzmann.
\newblock A survey of stroke-based rendering.
\newblock {\em IEEE Computer Graphics and Applications}, 23(4):70--81, 2003.

\bibitem{hertzmann2001image}
Aaron Hertzmann, Charles~E Jacobs, Nuria Oliver, Brian Curless, and David~H
  Salesin.
\newblock Image analogies.
\newblock In {\em Proceedings of the 28th Annual Conference on Computer
  Graphics and Interactive Techniques (SIGGRAPH)}, pages 327--340. ACM, 2001.

\bibitem{huang2017real}
Haozhi Huang, Hao Wang, Wenhan Luo, Lin Ma, Wenhao Jiang, Xiaolong Zhu, Zhifeng
  Li, and Wei Liu.
\newblock Real-time neural style transfer for videos.
\newblock In {\em IEEE Conference on Computer Vision and Pattern Recognition
  (CVPR)}, pages 783--791, 2017.

\bibitem{huang2011arcimboldo}
Hua Huang, Lei Zhang, and Hong-Chao Zhang.
\newblock Arcimboldo-like collage using internet images.
\newblock {\em ACM Transactions on Graphics (ToG)}, 30(6):155, 2011.

\bibitem{huang2017arbitrary}
Xun Huang and Serge Belongie.
\newblock Arbitrary style transfer in real-time with adaptive instance
  normalization.
\newblock In {\em IEEE International Conference on Computer Vision (ICCV)},
  pages 1501--1510, 2017.

\bibitem{jing2018stroke}
Yongcheng Jing, Yang Liu, Yezhou Yang, Zunlei Feng, Yizhou Yu, Dacheng Tao, and
  Mingli Song.
\newblock Stroke controllable fast style transfer with adaptive receptive
  fields.
\newblock In {\em European Conference on Computer Vision (ECCV)}, pages
  238--254, 2018.

\bibitem{jing2017neural}
Y. Jing, Yezhou Yang, Zunlei Feng, Jingwen Ye, Yizhou Yu, and Mingli Song.
\newblock Neural style transfer: A review.
\newblock {\em IEEE Transactions on Visualization and Computer Graphics}, 2019.

\bibitem{johnson2016perceptual}
Justin Johnson, Alexandre Alahi, and Li Fei-Fei.
\newblock Perceptual losses for real-time style transfer and super-resolution.
\newblock In {\em European Conference on Computer Vision (ECCV)}, pages
  694--711, 2016.

\bibitem{kato2018renderer}
Hiroharu Kato, Yoshitaka Ushiku, and Tatsuya Harada.
\newblock Neural 3d mesh renderer.
\newblock In {\em IEEE Conference on Computer Vision and Pattern Recognition
  (CVPR)}, 2018.

\bibitem{kingma2014adam}
Diederik~P. Kingma and Jimmy Ba.
\newblock Adam: A method for stochastic optimization.
\newblock {\em \tt arXiv:1412.6980[cs.LG]}, 2014.

\bibitem{kolkin2019style}
Nicholas Kolkin, Jason Salavon, and Gregory Shakhnarovich.
\newblock Style transfer by relaxed optimal transport and self-similarity.
\newblock In {\em IEEE Conference on Computer Vision and Pattern Recognition
  (CVPR)}, pages 10051--10060, 2019.

\bibitem{Kotovenko_2019_ICCV}
Dmytro Kotovenko, Artsiom Sanakoyeu, Sabine Lang, and Bj\"orn Ommer.
\newblock Content and style disentanglement for artistic style transfer.
\newblock In {\em IEEE International Conference on Computer Vision (ICCV)},
  2019.

\bibitem{kotovenko2019content}
Dmytro Kotovenko, Artsiom Sanakoyeu, Pingchuan Ma, Sabine Lang, and Bj\"orn
  Ommer.
\newblock A content transformation block for image style transfer.
\newblock In {\em IEEE Conference on Computer Vision and Pattern Recognition
  (CVPR)}, pages 10032--10041, 2019.

\bibitem{li2018carigan}
Wenbin Li, Wei Xiong, Haofu Liao, Jing Huo, Yang Gao, and Jiebo Luo.
\newblock Carigan: Caricature generation through weakly paired adversarial
  learning.
\newblock {\em \tt arXiv:1811.00445[cs.CV]}, 2018.

\bibitem{li2019learning}
Xueting Li, Sifei Liu, Jan Kautz, and Ming-Hsuan Yang.
\newblock Learning linear transformations for fast arbitrary style transfer.
\newblock In {\em IEEE Conference on Computer Vision and Pattern Recognition
  (CVPR)}, 2019.

\bibitem{li2017diversified}
Yijun Li, Chen Fang, Jimei Yang, Zhaowen Wang, Xin Lu, and Ming-Hsuan Yang.
\newblock Diversified texture synthesis with feed-forward networks.
\newblock In {\em IEEE Conference on Computer Vision and Pattern Recognition
  (CVPR)}, pages 3920--3928, 2017.

\bibitem{li2017universal}
Yijun Li, Chen Fang, Jimei Yang, Zhaowen Wang, Xin Lu, and Ming-Hsuan Yang.
\newblock Universal style transfer via feature transforms.
\newblock In {\em Advances in Neural Information Processing Systems (NIPS)},
  pages 386--396, 2017.

\bibitem{li2018closed}
Y. Li, Ming-Yu Liu, Xueting Li, Ming-Hsuan Yang, and Jan Kautz.
\newblock A closed-form solution to photorealistic image stylization.
\newblock In {\em European Conference on Computer Vision (ECCV)}, 2018.

\bibitem{Liao:2017:VAT:3072959.3073683}
Jing Liao, Yuan Yao, Lu Yuan, Gang Hua, and Sing~Bing Kang.
\newblock Visual attribute transfer through deep image analogy.
\newblock {\em ACM Transactions on Graphics (ToG)}, 36(4):120:1--120:15, 2017.

\bibitem{lin2014microsoft}
Tsung-Yi Lin, Michael Maire, Serge Belongie, James Hays, Pietro Perona, Deva
  Ramanan, Piotr Doll{\'a}r, and C~Lawrence Zitnick.
\newblock Microsoft coco: Common objects in context.
\newblock In {\em European Conference on Computer Vision (ECCV)}, pages
  740--755, 2014.

\bibitem{luan2017deep}
Fujun Luan, Sylvain Paris, Eli Shechtman, and Kavita Bala.
\newblock Deep photo style transfer.
\newblock In {\em IEEE Conference on Computer Vision and Pattern Recognition
  (CVPR)}, pages 6997--7005, 2017.

\bibitem{luo2016understanding}
Wenjie Luo, Yujia Li, Raquel Urtasun, and Richard Zemel.
\newblock Understanding the effective receptive field in deep convolutional
  neural networks.
\newblock In {\em Advances in Neural Information Processing Systems (NIPS)},
  pages 4898--4906, 2016.

\bibitem{mechrez2017photorealistic}
Roey Mechrez, Eli Shechtman, and Lihi Zelnik-Manor.
\newblock Photorealistic style transfer with screened poisson equation.
\newblock In {\em British Machine Vision Conference (BMVC)}, 2017.

\bibitem{mechrez2018contextual}
Roey Mechrez, Itamar Talmi, and Lihi Zelnik-Manor.
\newblock The contextual loss for image transformation with non-aligned data.
\newblock In {\em European Conference on Computer Vision (ECCV)}, pages
  768--783, 2018.

\bibitem{paszke2017automatic}
Adam Paszke, Sam Gross, Soumith Chintala, Gregory Chanan, Edward Yang, Zachary
  DeVito, Zeming Lin, Alban Desmaison, Luca Antiga, and Adam Lerer.
\newblock Automatic differentiation in {PyTorch}.
\newblock In {\em NIPS Autodiff Workshop}, 2017.

\bibitem{Ruder2018ArtisticST}
Manuel Ruder, Alexey Dosovitskiy, and Thomas Brox.
\newblock Artistic style transfer for videos and spherical images.
\newblock {\em International Journal of Computer Vision}, 126:1199--1219, 2018.

\bibitem{selim2016painting}
Ahmed Selim, Mohamed Elgharib, and Linda Doyle.
\newblock Painting style transfer for head portraits using convolutional neural
  networks.
\newblock {\em ACM Transactions on Graphics (ToG)}, 35(4):129, 2016.

\bibitem{shi2019warpgan}
Yichun Shi, Debayan Deb, and Anil~K Jain.
\newblock Warpgan: Automatic caricature generation.
\newblock In {\em IEEE Conference on Computer Vision and Pattern Recognition
  (CVPR)}, pages 10762--10771, 2019.

\bibitem{simonyan2014very}
Karen Simonyan and Andrew Zisserman.
\newblock Very deep convolutional networks for large-scale image recognition.
\newblock In {\em The International Conference on Learning Representations
  (ICLR)}, 2015.

\bibitem{Snelgrove:2017}
Xavier Snelgrove.
\newblock High-resolution multi-scale neural texture synthesis.
\newblock In {\em SIGGRAPH ASIA 2017 Technical Briefs}. ACM, 2017.

\bibitem{willats1997art}
John Willats.
\newblock {\em Art and representation: New principles in the analysis of
  pictures}.
\newblock Princeton University Press, 1997.

\bibitem{xu2018learning}
Zheng Xu, Michael Wilber, Chen Fang, Aaron Hertzmann, and Hailin Jin.
\newblock Learning from multi-domain artistic images for arbitrary style
  transfer.
\newblock {\em \tt arXiv:1805.09987[cs.CV]}, 2018.

\bibitem{Yang_textNST_2019_ICCV}
Shuai Yang, Zhangyang Wang, Zhaowen Wang, Ning Xu, Jiaying Liu, and Zongming
  Guo.
\newblock Controllable artistic text style transfer via shape-matching gan.
\newblock In {\em IEEE International Conference on Computer Vision (ICCV)},
  2019.

\bibitem{Yaniv2019}
Jordan Yaniv, Yael Newman, and Ariel Shamir.
\newblock The face of art: Landmark detection and geometric style in portraits.
\newblock {\em ACM Transactions on Graphics (ToG)}, 38(4):60, 2019.

\bibitem{yao2019attention}
Yuan Yao, Jianqiang Ren, Xuansong Xie, Weidong Liu, Yong-Jin Liu, and Jun Wang.
\newblock Attention-aware multi-stroke style transfer.
\newblock In {\em IEEE Conference on Computer Vision and Pattern Recognition
  (CVPR)}, 2019.

\bibitem{Yoo_2019_ICCV}
Jaejun Yoo, Youngjung Uh, Sanghyuk Chun, Byeongkyu Kang, and Jung-Woo Ha.
\newblock Photorealistic style transfer via wavelet transforms.
\newblock In {\em IEEE International Conference on Computer Vision (ICCV)},
  2019.

\bibitem{yu2004framework}
Jingyi Yu and Leonard McMillan.
\newblock A framework for multiperspective rendering.
\newblock In {\em Rendering Techniques}, pages 61--68, 2004.

\bibitem{yuphoto}
Ziqiang Zheng, Wang Chao, Zhibin Yu, Nan Wang, Haiyong Zheng, and Bing Zheng.
\newblock Unpaired photo-to-caricature translation on faces in the wild.
\newblock {\em Neurocomputing}, 2019.

\end{thebibliography}
}

\end{document}